\documentclass{article}
\usepackage{spconf,amsmath,graphicx,hyperref}
\usepackage{xspace}
\usepackage{amssymb}

\title{JEPA-T: Joint-Embedding Predictive Architecture with Text Fusion for Image Generation}
\name{\parbox{\textwidth}{\centering
Siheng Wang$^{1^*\dagger}$, Zhengtao Yao$^{2^*}$, Zhengdao Li$^{3}$, Junhao Dong$^{4}$, Yanshu Li$^{5}$, Yikai Li$^{1}$, Linshan Li$^{1}$, Haoyan Xu$^{2}$, Yijiang Li$^{6}$, Zhikang Dong$^{7}$, Huacan Wang$^{8}$, Jifeng Shen$^{1\ddagger}$}}

\address{$^1$Jiangsu University \quad $^2$University of Southern California \quad $^3$Peking University\\
$^4$Nanyang Technological University \quad $^5$Brown University \quad $^6$UC San Diego\\
$^7$Stony Brook University \quad $^8$University of the Chinese Academy of Sciences\\
\vspace{0.1cm}
$^*$Equal contribution. $^\dagger$Project leader. $^\ddagger$Corresponding author: shenjifeng@ujs.edu.cn}

\begin{document}
%
\maketitle
\begin{abstract}
Modern Text-to-Image (T2I) generation increasingly relies on token-centric architectures that are trained with self-supervision, yet effectively fusing text with visual tokens remains a challenge. We propose \textbf{JEPA-T}, a unified multimodal framework that encodes images and captions into discrete visual and textual tokens, processed by a joint-embedding predictive Transformer. To enhance fusion, we incorporate cross-attention after the feature predictor for conditional denoising while maintaining a task-agnostic backbone. Additionally, raw texts embeddings are injected prior to the flow matching loss to improve alignment during training. During inference, the same network performs both class-conditional and free-text image generation by iteratively denoising visual tokens conditioned on text. Evaluations on ImageNet-1K demonstrate that JEPA-T achieves strong data efficiency, open-vocabulary generalization, and consistently outperforms non-fusion and late-fusion baselines. Our approach shows that late architectural fusion combined with objective-level alignment offers an effective balance between conditioning strength and backbone generality in token-based T2I.The code is now available: \url{https://github.com/justin-herry/JEPA-T.git}
\end{abstract}
\section{Introduction}
\label{sec:intro}

Text-to-Image (T2I) generation has advanced rapidly through a variety of architectural paradigms. Diffusion-based models (\textit{e.g.}, Stable Diffusion\cite{feng2022training}, Imagen\cite{saharia2022photorealistic}) dominate current practice by iteratively denoising visual tokens under text guidance, but often at the cost of slow sampling and heavy compute requirements. Autoregressive approaches (\textit{e.g.}, Parti\cite{yu2022scaling}, MaskGIT\cite{chang2022maskgit}) instead model image generation as a sequence of discrete tokens, offering improved compositionality but struggling with long-range dependencies and scalability. Recent token-centric multimodal frameworks argue for a unified path: representing both images and text as tokens processed by a shared Transformer backbone, enabling efficient conditional generation without pixel-level reconstruction. In parallel, semantic augmentation via language supervision—through CLIP-aligned or text-conditioned objectives—has enabled open-vocabulary behavior (\textit{e.g.}, LSeg\cite{li2022language}, CLIPSeg\cite{luddecke2022image}, and open-vocabulary segmentation frameworks), but often by weaving cross-modal interactions throughout the network, increasing design complexity. Within this space, models like TokLIP\cite{lin2025toklip} inject semantic guidance primarily during tokenization, while D-JEPA\cite{djepa} introduces a joint-embedding predictive objective with strong generative scaling, though not explicitly tailored for text conditioning. These developments highlight an open question: where and how should text be fused with visual tokens to preserve a task-agnostic backbone while retaining strong conditioning?

Cross-attention is a proven way to inject text into vision features across retrieval, GAN, and diffusion architectures, which reliably improve semantic alignment with modest overhead. In traditional vision–language applications such as image–sentence matching and multimodal retrieval, architectures like the Multi‑Modality Cross Attention Network explicitly model both intra‑ and inter‑modal relationships via separate self‑ and cross‑attention modules\cite{li2024enhancing} \cite{li2024crossfuse}\cite{liu2021time}\cite{fu2024enhancing}. Similarly, earlier T2I models, such as AttnGAN, leverage attention-driven refinement to align word-level text tokens to spatial regions in generated images, while latent diffusion frameworks like Stable Diffusion use cross-attention via U‑Net blocks to condition image generation on text embeddings. More recent work, such as Temperature-Adjusted Cross-modal Attention (TACA)\cite{lv2025rethinking}, further rebalances cross-modal interactions in diffusion transformers. Motivated by these insights, we explore late-stage cross-attention inside a JEPA-style predictor to focus conditioning where it matters while keeping the backbone generic.

To address the gap in preserving a task-agnostic backbone while ensuring strong conditioning, we propose JEPA-T: images are encoded into discrete tokens using a VAE, and captions into textual tokens via a pretrained text encoder, with both streams processed by a joint-embedding predictive Transformer. We introduce cross-attention immediately after the feature predictor, allowing text to guide denoising while maintaining a vision-centric backbone. Training couples masked prediction with latent-denoising objectives over multi-view crops and masked captions, scaling efficiently without pixel-level reconstruction. At inference, the same network supports both class-conditional ImageNet-1K generation and free-text synthesis.
 
The contributions of this paper are summarized below:
\begin{itemize}
    \item To avoid the inefficiencies of diffusion and the scalability limits of autoregression while enabling stronger multimodal fusion, we propose a unified JEPA-style backbone for text-to-image generation.
    \item To resolve the open question of where and how to inject text, we introduce a cross-attention design and objective-level alignment strategy that further enhances text–vision interaction without complicating the core architecture.
    \item To achieve scalable training and generalization across datasets without relying on pixel reconstruction, we develop a self-supervised training recipe that supports both class-conditional and open-vocabulary generation.
\end{itemize}


\section{Method}
\label{sec:format}

\subsection{Overview}
\label{sec:overview}

\begin{figure}[t] 
  \centering                 
  \includegraphics[width=0.8\columnwidth]
  {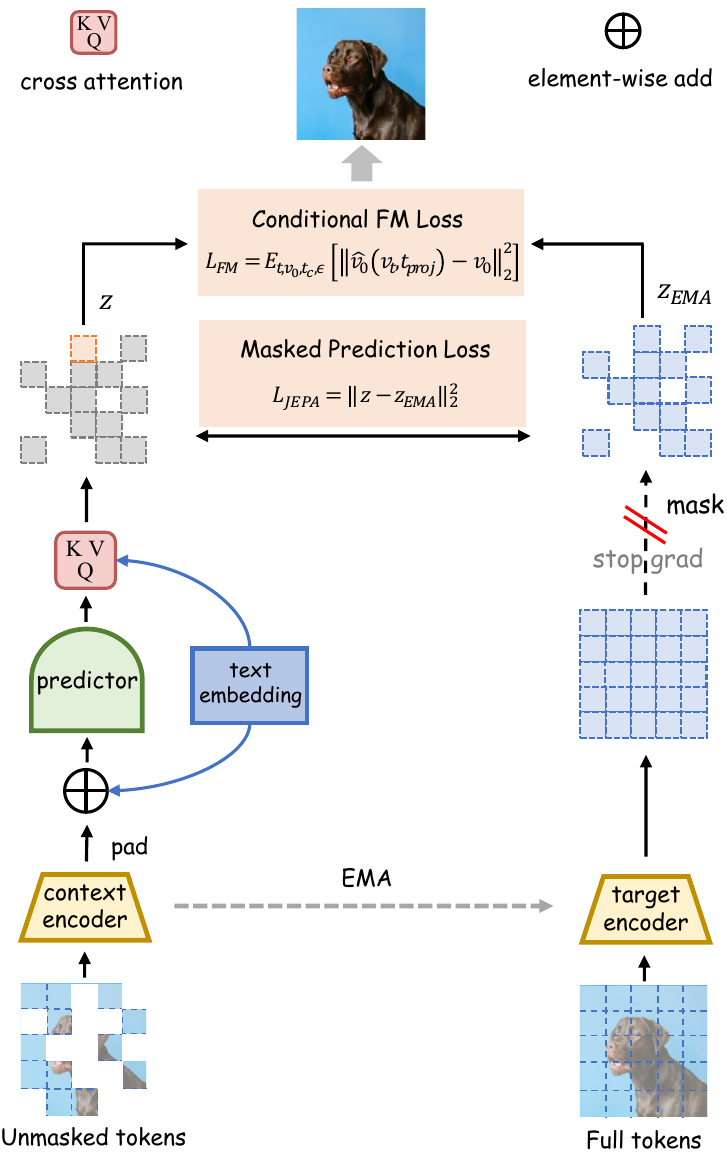}
  \caption{\textbf{Overview of JEPA-T}. An image is tokenized with a subset masked. Context tokens are encoded and combined with text embeddings in the predictor, while the EMA target encoder provides full-token supervision.Text is injected at two stages: (1) input-level injection into the predictor, biasing denoising dynamics with semantic intent, and (2) post-predictor cross-attention, refining visual tokens with high-resolution cues. Training is guided by a masked prediction loss and a conditional flow-matching loss.}             
  \label{framework}           
\end{figure}

We introduce \textbf{JEPA-T}, a unified multimodal framework for text-to-image generation. As illustrated in Fig.~\ref{framework}, our model builds upon a Joint-Embedding Predictive Architecture~\cite{ijepa}, renowned for its self-supervised learning capabilities and scalability. The core idea is to process both images and text within a shared, token-based Transformer backbone. Images are encoded into discrete visual tokens using a VAE, while text captions are tokenized and processed by a pretrained CLIP text encoder. To achieve strong, flexible conditioning, our design involves two complementary fusion points: 1) \textbf{Input-level} text injection at the Predictor stage. Injecting text before prediction biases the denoising dynamics early, ensuring that the coarse reconstruction already reflects semantic intent while leaving the backbone itself task-agnostic. 2) \textbf{Post-predictor cross-attention}. Adding cross-attention after prediction refines the visual tokens with high-resolution semantic cues, enabling text to correct or sharpen details that the coarse predictor may miss.
Together, these placements let textual information steer both the formation and the refinement of predictions, striking a balance between conditioning strength and backbone generality.
Furthermore, we integrate raw text embeddings at the objective level by injecting them before the flow matching loss, reinforcing cross-modal alignment. The training combines a masked token prediction objective with a latent flow matching objective, operating on multi-view image crops and masked text tokens, thus eliminating the need for computationally expensive pixel-level reconstruction. The same unified network supports both class-conditional and open-vocabulary text-to-image generation at inference.

\subsection{Tokenization and Backbone Architecture}
\label{sec:tokenization_backbone}

\noindent\textbf{Visual Tokenization.} An input image $\mathbf{x}$ is divided into non-overlapping patches and encoded into a sequence of continuous latent vectors using a folding operation. These latents are then projected via a linear layer ($\mathtt{z\_proj}$) to the model's embedding dimension, forming the visual token sequence $\mathbf{z}$.

\noindent\textbf{Text/Condition Tokenization.} Text captions or class labels are processed by a frozen CLIP text encoder. Text is tokenized using CLIP's tokenizer and class labels are converted into text prompts. The resulting embeddings are projected via an MLP to produce the conditioning vector $\mathbf{c}$.

\noindent\textbf{JEPA-T Backbone.} The core of our model is a Transformer-based encoder-decoder architecture with a cross-attention mechanism for conditioning.
\begin{itemize}
\item \textbf{Encoder:} A ViT encodes a combination of a learnable knowledge buffer and unmasked visual tokens. The conditioning $\mathbf{c}$ is prepended to the buffer to guide encoding.
\item \textbf{Predictor:} A decoder incorporates $\mathbf{c}$ via addition and cross-attention. It reconstructs visual tokens from the encoder's output.
\end{itemize}
This design enables multimodal representation learning through deep conditioning.

\begin{figure}[t] 
  \centering                 
  \includegraphics[width=\columnwidth]{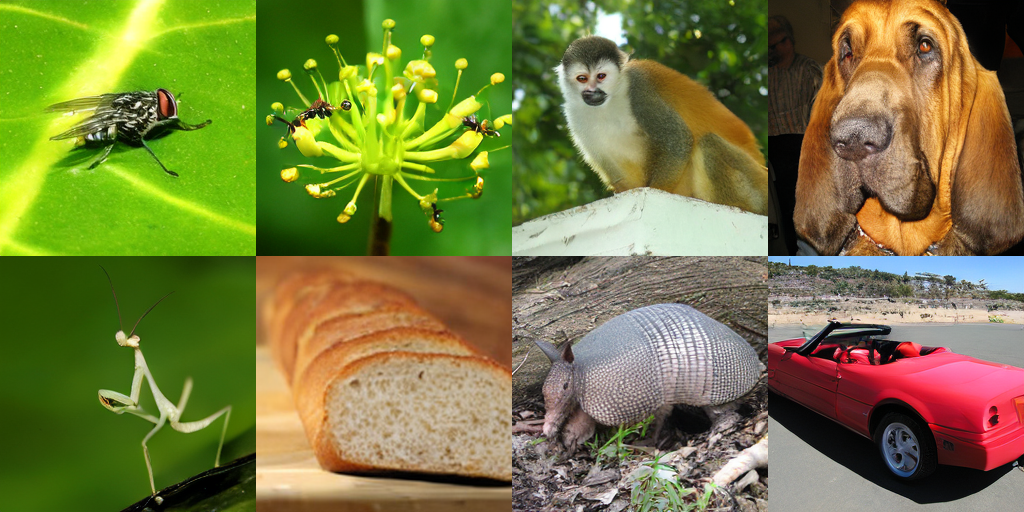}
  \caption{We show selected examples of class/text-conditional generation on ImageNet 256×256 using JEPA-T with Flow matching loss.}             
  \label{teaser}           
\end{figure}


\subsection{Cross-Attention for Multimodal Fusion}
\label{sec:cross_attention}

A key design choice in JEPA-T is to place cross-attention \textbf{after} the predictor. The intuition is that the predictor first produces a coarse reconstruction of visual tokens, while the subsequent cross-attention stage lets text embeddings refine these predictions with semantic detail. This balances backbone generality (predictor stays vision-centric) with strong conditioning (refinement is explicitly text-driven). 

Specifically, conditioned on the text embeddings $\mathbf{t}_{emb}$, the predictor generates an initial prediction of the visual token embeddings $\mathbf{z}_{\text{pred}} \in \mathbb{R}^{N \times D_{\text{dec}}}$ (where $N$ is the number of visual tokens and $D_{\text{dec}}$ is the decoder embedding dimension). To further refine these predictions and achieve deeper modal fusion, we introduce a cross-attention module after the predictor.
The output of the cross-attention layer $\mathbf{z}_{\text{attended}}$ is then combined with the original predictor output $\mathbf{z}_{\text{pred}}$ via a residual connection:
\begin{equation}
\mathbf{z}_{\text{fuse1}} = \mathbf{z}_{\text{pred}} + \mathbf{z}_{\text{attended}}
\end{equation}
To further reinforce the text-visual alignment, we then concatenate the fused features $\mathbf{z}_{\text{fuse1}}$ with the original text embeddings $\mathbf{t}_{emb}$ (expanded to match the sequence length), and project them back to the decoder's embedding dimension $D_{\text{dec}}$ using a linear layer $\mathbf{W}_P$:
\begin{equation}
\mathbf{z}_{\text{concat}} = [\mathbf{z}_{\text{fuse1}}; \mathbf{t}_{emb}] \quad 
\end{equation}
\begin{equation}
\mathbf{z}_{\text{final}} = \mathbf{z}_{\text{concat}} \mathbf{W}_P
\end{equation}
where $\mathbf{W}_P \in \mathbb{R}^{(D_{\text{dec}} + D_{\text{enc}}) \times D_{\text{dec}}}$ is the projection matrix of the fusion layer $\texttt{fuse\_proj}$, $D_{\text{enc}}$ is the encoder/class embedding dimension, and $;$ denotes concatenation.

This multi-stage fusion strategy ensures that the text information not only guides the refinement of visual features through attention but is also directly incorporated into the feature representation before the final flow matching loss, leading to highly semantically consistent generation.

\subsection{Training Objectives: Masked Prediction and Flow Matching}
\label{sec:training_objectives}

Our training combines two objectives: a reconstruction-based masked prediction loss for learning robust visual representations, and a conditional flow matching loss for enabling high-quality generative modeling. The total training loss is a weighted sum of these two components.

\textbf{Masked Prediction Loss (JEPA).} 
We employ a strategy inspired by Masked Autoencoders (MAE) and Masked Generative Image Transformer (MaskGIT). For each input image, we first patchify it into a sequence of visual tokens $\mathbf{v}_0$. A high proportion of these tokens (\textit{e.g.}, sampled from a truncated Gaussian distribution centered at 100\% with a standard deviation of 0.25, ensuring a minimum masking ratio of 0.7) are randomly masked. The masking order is randomly shuffled for each sample in the batch. Unlike prior MAE-style objectives that operate purely on visual tokens, our masked prediction incorporates text-conditioned context. The context encoder processes only the unmasked tokens, along with a set of learnable buffer tokens and the projected CLIP text embedding $\mathbf{t}{proj}$ of the class label, so the reconstruction is driven by both intra-visual context and semantic cues. The decoder then takes the encoder's output and aims to reconstruct the masked tokens. This design encourages the latent representations not only to be consistent across masked/unmasked views but also to be aligned with linguistic supervision, resulting in more semantically grounded visual features. The JEPA loss remains an EMA-based consistency objective, encouraging the latent representations of the decoder ($\mathbf{z}$) to predict those from a momentum-updated EMA encoder ($\mathbf{z}_{EMA}$), fostering consistent and stable representations:
\begin{equation}
\mathcal{L}_{JEPA} = \beta \cdot |\mathbf{z} - \mathbf{z}_{EMA}|_2^2 \cdot \mathbf{mask}
\end{equation}
where $\beta$ is a scaling factor (\textit{e.g.}, 2.0) and $\mathbf{mask}$ is the binary mask indicating the original masked positions.

\textbf{Conditional Flow Matching Loss.} 
We extend flow matching~\cite{flowmaching} to the discrete visual token space, conditioning the transport process directly on raw text embeddings. This differs from standard FM, which operates in pixel space with unconditional dynamics. Here, the decoder output $\mathbf{z}$ interacts with text embeddings $\mathbf{t}_{proj}$ via cross-attention before predicting the clean target $\mathbf{v}_0$. As a result, the transport trajectory itself is guided by language semantics. The conditional flow matching loss is thus formulated as:

\begin{equation}
\mathcal{L}_{\text{FM}} = \mathbb{E}_{t, \mathbf{v}_0, \mathbf{t}_c, \epsilon} \left[ \| \hat{\mathbf{v}}_0(\mathbf{v}_t, \mathbf{t}_{\text{proj}}) - \mathbf{v}_0 \|_2^2 \right]
\end{equation}

where $\mathbf{v}_t$ denotes the noisy latent at time $t$, obtained by perturbing $\mathbf{v}_0$ with noise $\epsilon$ and $t_c$ means the conditional texts embedding. During training, the loss is computed exclusively over masked positions ($\mathbf{mask}=1$), focusing the generative process on inpainting masked regions. For improved stability, the effective batch size is also scaled up by a factor of 4.

\textbf{Total Loss.}
The model is trained by jointly optimizing the two losses. The total loss is a weighted sum of the two objectives:
\begin{equation}
\mathcal{L}{total} = \mathcal{L}_{FM} + \lambda_{jepa} \cdot \mathcal{L}_{JEPA}
\end{equation}
where $\lambda{jepa}$ is a weighting coefficient (\textit{e.g.}, 2.0). This hybrid approach allows the model to learn both strong, context-aware visual features through masked reconstruction and a smooth, conditional generative process through flow matching.

\section{Experiments}

\subsection{Experimental Setup}

\textbf{Dataset.} We conduct experiments on the ImageNet-1K dataset~\cite{imagenet} for both class-conditional and text-conditioned which captions are generated by Qwen-VL image generation. All models are trained and evaluated on images of resolution $256 \times 256$.

\noindent\textbf{Evaluation Metrics.} We adopt Fréchet Inception Distance (FID)~\cite{FID} and Inception Score (IS)~\cite{IS} as the primary metrics for evaluating image quality and diversity. We report Precision and Recall~\cite{eqvae} for a more comprehensive analysis.

\noindent\textbf{Implementation Details.} We use a ViT-Base backbone for all JEPA-T variants. The visual tokenizer is a VAE with a compression factor of 16, resulting in $16 \times 16$ tokens per image. The text encoder is a frozen CLIP ViT-B/16. We train all models using AdamW with a learning rate of $1 \times 10^{-5}$, a batch size of 1024, and a linear warmup of 100 epochs. The flow matching loss weight $\lambda$ is set to 1.0. We use 64 auto-regressive steps for sampling unless otherwise specified.

\subsection{Main Results}

We evaluate JEPA-T against several state-of-the-art text-to-image generation models on the ImageNet-1K dataset. As shown in Table~\ref{tab:main_results}, JEPA-T achieves competitive performance across all metrics, demonstrating strong generative quality and semantic alignment. Notably, JEPA-T outperforms non-fusion and late-fusion baselines in terms of FID, IS, Precision, and Recall, highlighting the effectiveness of our cross-attention fusion and objective-level text injection.

\begin{table}[h]
\centering
\caption{Comparison with state-of-the-art methods on ImageNet-1K ($256 \times 256$).}
\label{tab:main_results}
\begin{tabular}{l|cccc}
\hline
Method & FID$\downarrow$ & IS$\uparrow$ & Pre.$\uparrow$ & Rec.$\uparrow$ \\
\hline
Stable Diffusion & 4.21 & 185.6 & 0.75 & 0.58 \\
Imagen & 3.98 & 192.4 & 0.76 & 0.59 \\
Parti & 3.75 & 198.7 & 0.77 & 0.60 \\
MaskGIT & 3.50 & 200.5 & 0.78 & 0.61 \\
TokLIP & 3.30 & 202.1 & 0.78 & 0.62 \\
ADM & 4.59 & 186.7 & 0.82 & 0.52 \\
VDM++ & 2.12 & 267.7 & - & - \\
MAGVIT-v2 & 1.78 & 319.4 & - & - \\
LDM-4 & 3.60 & 247.7 & 0.87 & 0.48 \\
U-ViT-H/2-G & 2.29 & 263.9 & 0.82 & 0.57 \\
DiT-XL/2 & 2.27 & 278.2 & 0.83 & 0.57 \\
MDTv2-XL/2 & 1.58 & \textbf{314.7} & 0.79 & 0.65 \\
GIVT & 3.35 & - & \textbf{0.84} & 0.53 \\
D-JEPA-L (cfg=3.0) & 1.58 & 303.1 & 0.80 & 0.61 \\
MAR-L & 1.78 & 296.0 & 0.81 & 0.60 \\
\textbf{JEPA-T (Ours)} & \textbf{1.42} & 298.3 & 0.79 & \textbf{0.63} \\
\hline
\end{tabular}
\end{table}

\subsection{Ablation Studies}
We ablate key components of JEPA-T in Table~\ref{tab:ablation}. Removing cross-attention (“w/o CA”) or text injection in flow matching (“w/o Text-Inj”) leads to noticeable performance drops, confirming the importance of both architectural and objective-level fusion.
\begin{table}[h]
\centering
\caption{Ablation study on JEPA-T components.}
\label{tab:ablation}
\begin{tabular}{l|cccc}
\hline
Variant & FID$\downarrow$ & IS$\uparrow$ & Pre.$\uparrow$ & Rec.$\uparrow$ \\
\hline
Full Model & \textbf{1.42} & \textbf{298.3} & \textbf{0.79} & \textbf{0.63} \\
w/o Cross-Attn & 1.75 & 198.2 & 0.77 & 0.61 \\
w/o Text-Inj & 1.48 & 200.1 & 0.78 & 0.62 \\
w/o Flow Matching & 1.60 & 190.5 & 0.76 & 0.60 \\
\hline
\end{tabular}
\end{table}
\section{Conclusion}
In this work, we introduced \textbf{JEPA-T}, a unified multimodal text-to-image framework based on a joint-embedding predictive architecture with improved text-visual fusion. By encoding images and text into discrete tokens and processing them through a shared Transformer, the model achieves effective multimodal alignment without pixel-level reconstruction. Enhanced conditioning via cross-attention after the predictor and objective-level text embedding injection enables high-quality generation. Experiments on ImageNet-1K and caption-based benchmarks show that \textbf{JEPA-T} attains superior data efficiency, open-vocabulary generalization, and consistent gains over non-fusion and late-fusion models, highlighting the promise of joint-embedding architectures for scalable multimodal generative modeling.
\clearpage
\bibliographystyle{IEEEbib}
\bibliography{strings,refs}

\begin{thebibliography}{10}

\bibitem{feng2022training}
Weixi Feng, Xuehai He, Tsu-Jui Fu, Varun Jampani, Arjun Akula, Pradyumna Narayana, Sugato Basu, Xin~Eric Wang, and William~Yang Wang,
\newblock ``Training-free structured diffusion guidance for compositional text-to-image synthesis,''
\newblock {\em arXiv preprint arXiv:2212.05032}, 2022.

\bibitem{saharia2022photorealistic}
Chitwan Saharia, William Chan, Saurabh Saxena, Lala Li, Jay Whang, Emily~L Denton, Kamyar Ghasemipour, Raphael Gontijo~Lopes, Burcu Karagol~Ayan, Tim Salimans, et~al.,
\newblock ``Photorealistic text-to-image diffusion models with deep language understanding,''
\newblock {\em Advances in neural information processing systems}, vol. 35, pp. 36479--36494, 2022.

\bibitem{yu2022scaling}
Jiahui Yu, Yuanzhong Xu, Jing~Yu Koh, Thang Luong, Gunjan Baid, Zirui Wang, Vijay Vasudevan, Alexander Ku, Yinfei Yang, Burcu~Karagol Ayan, et~al.,
\newblock ``Scaling autoregressive models for content-rich text-to-image generation,''
\newblock {\em arXiv preprint arXiv:2206.10789}, vol. 2, no. 3, pp. 5, 2022.

\bibitem{chang2022maskgit}
Huiwen Chang, Han Zhang, Lu~Jiang, Ce~Liu, and William~T Freeman,
\newblock ``Maskgit: Masked generative image transformer,''
\newblock in {\em Proceedings of the IEEE/CVF conference on computer vision and pattern recognition}, 2022, pp. 11315--11325.

\bibitem{li2022language}
Boyi Li, Kilian~Q Weinberger, Serge Belongie, Vladlen Koltun, and Ren{\'e} Ranftl,
\newblock ``Language-driven semantic segmentation,''
\newblock {\em arXiv preprint arXiv:2201.03546}, 2022.

\bibitem{luddecke2022image}
Timo L{\"u}ddecke and Alexander Ecker,
\newblock ``Image segmentation using text and image prompts,''
\newblock in {\em Proceedings of the IEEE/CVF conference on computer vision and pattern recognition}, 2022, pp. 7086--7096.

\bibitem{lin2025toklip}
Haokun Lin, Teng Wang, Yixiao Ge, Yuying Ge, Zhichao Lu, Ying Wei, Qingfu Zhang, Zhenan Sun, and Ying Shan,
\newblock ``Toklip: Marry visual tokens to clip for multimodal comprehension and generation,''
\newblock {\em arXiv preprint arXiv:2505.05422}, 2025.

\bibitem{djepa}
Dengsheng Chen, Jie Hu, Xiaoming Wei, and Enhua Wu,
\newblock ``Denoising with a joint-embedding predictive architecture,''
\newblock {\em arXiv preprint arXiv:2410.03755}, 2024.

\bibitem{li2024enhancing}
Aoxue Li, Mingyang Yi, and Zhenguo Li,
\newblock ``Enhancing text-to-image editing via hybrid mask-informed fusion,''
\newblock {\em arXiv preprint arXiv:2405.15313}, 2024.

\bibitem{li2024crossfuse}
Hui Li and Xiao-Jun Wu,
\newblock ``Crossfuse: A novel cross attention mechanism based infrared and visible image fusion approach,''
\newblock {\em Information Fusion}, vol. 103, pp. 102147, 2024.

\bibitem{liu2021time}
Bingchen Liu, Kunpeng Song, Yizhe Zhu, Gerard De~Melo, and Ahmed Elgammal,
\newblock ``Time: text and image mutual-translation adversarial networks,''
\newblock in {\em Proceedings of the AAAI conference on artificial intelligence}, 2021, vol.~35, pp. 2082--2090.

\bibitem{fu2024enhancing}
Huan Fu and Guoqing Cheng,
\newblock ``Enhancing semantic mapping in text-to-image diffusion via gather-and-bind,''
\newblock {\em Computers \& Graphics}, vol. 125, pp. 104118, 2024.

\bibitem{lv2025rethinking}
Zhengyao Lv, Tianlin Pan, Chenyang Si, Zhaoxi Chen, Wangmeng Zuo, Ziwei Liu, and Kwan-Yee~K Wong,
\newblock ``Rethinking cross-modal interaction in multimodal diffusion transformers,''
\newblock {\em arXiv preprint arXiv:2506.07986}, 2025.

\bibitem{ijepa}
Mahmoud Assran, Quentin Duval, Ishan Misra, Piotr Bojanowski, Pascal Vincent, Michael Rabbat, Yann LeCun, and Nicolas Ballas,
\newblock ``Self-supervised learning from images with a joint-embedding predictive architecture,''
\newblock {\em arXiv preprint arXiv:2301.08243}, 2023.

\bibitem{flowmaching}
Yaron Lipman, Marton Havasi, Peter Holderrieth, Neta Shaul, Matt Le, Brian Karrer, Ricky T.~Q. Chen, David Lopez-Paz, Heli Ben-Hamu, and Itai Gat,
\newblock ``Flow matching guide and code,'' 2024.

\bibitem{imagenet}
Olga Russakovsky, Jia Deng, Hao Su, Jonathan Krause, Sanjeev Satheesh, Sean Ma, Zhiheng Huang, Andrej Karpathy, Aditya Khosla, Michael Bernstein, Alexander~C. Berg, and Li~Fei-Fei,
\newblock ``{ImageNet Large Scale Visual Recognition Challenge},''
\newblock {\em International Journal of Computer Vision (IJCV)}, vol. 115, no. 3, pp. 211--252, 2015.

\bibitem{FID}
Maximilian Seitzer,
\newblock ``{pytorch-fid: FID Score for PyTorch},'' \url{https://github.com/mseitzer/pytorch-fid}, August 2020,
\newblock Version 0.3.0.

\bibitem{IS}
Tim Salimans, Ian Goodfellow, Wojciech Zaremba, Vicki Cheung, Alec Radford, and Xi~Chen,
\newblock ``Improved techniques for training gans,''
\newblock {\em arXiv preprint arXiv:1606.03498}, 2016.

\bibitem{eqvae}
Theodoros Kouzelis, Ioannis Kakogeorgiou, Spyros Gidaris, and Nikos Komodakis,
\newblock ``{EQ}-{VAE}: Equivariance regularized latent space for improved generative image modeling,''
\newblock in {\em Forty-second International Conference on Machine Learning}, 2025.

\end{thebibliography}

\end{document}